\documentclass[10pt,twocolumn,letterpaper]{article}

\usepackage{wacv}
\usepackage{times}
\usepackage{epsfig}
\usepackage{graphicx}
\usepackage{amsmath}
\usepackage{amssymb}
\usepackage{subcaption}
\usepackage{relsize}
\usepackage{caption}
\usepackage{multirow}
\usepackage{makecell}
\newcolumntype{L}[1]{>{\raggedright\let\newline\\\arraybackslash\hspace{0pt}}m{#1}}
\newcolumntype{C}[1]{>{\centering\let\newline\\\arraybackslash\hspace{0pt}}m{#1}}
\newcolumntype{R}[1]{>{\raggedleft\let\newline\\\arraybackslash\hspace{0pt}}m{#1}}

\DeclareMathOperator*{\argmax}{argmax} 
\usepackage{multirow}
\usepackage{enumitem}
\newcommand{\comb}[2]{{}_{#1}\mathrm{C}_{#2}}

\usepackage{bm}
\DeclareMathOperator*{\argmin}{argmin} 

\wacvfinalcopy 

\def\wacvPaperID{****} 

\ifwacvfinal\fi
\begin{document}

\title{Learning to Align Multi-Camera Domains using Part-Aware Clustering \\ for Unsupervised Video  Person Re-Identification}

\author{
Youngeun Kim\\
KAIST\\
\and
Seokeon Choi\\
KAIST\\
\and
Taekyung Kim\\
KAIST\\
\and
Sumin Lee\\
KAIST\\
\and
Changick Kim\\
KAIST\\
\and
{\tt\small \{youngeunkim, seokeon, tkkim93, suminlee94, changick\}@kaist.ac.kr}
}
\maketitle

\begin{abstract}
Most video person re-identification (re-ID) methods are mainly based on supervised learning, which requires cross-camera ID labeling.
Since the cost of labeling increases dramatically as the number of cameras increases, it is difficult to apply the re-identification algorithm to a large camera network.
In this paper, we address the scalability issue by presenting deep representation learning without ID information across multiple cameras.
Technically, we train neural networks to generate both ID-discriminative and camera-invariant features.
To achieve the ID discrimination ability of the embedding features, we maximize feature distances between different person IDs within a camera by using a metric learning approach.
At the same time, considering each camera as a different domain, we apply adversarial learning across multiple camera domains for generating camera-invariant features.
We also propose a part-aware adaptation module, which effectively performs multi-camera domain invariant feature learning in different spatial regions.
We carry out comprehensive experiments on three public re-ID datasets (i.e., PRID-2011, iLIDS-VID, and MARS). 
Our method outperforms state-of-the-art methods by a large margin of about 20\% in terms of rank-1 accuracy on the large-scale MARS dataset.
\end{abstract}

\vspace{-2mm}

\section{Introduction}

Person re-identification (re-ID) \cite{imgsup_CVPR2018, BC1_ICCV2017, BC2_ECCV2018, PUL_TMM2018,   part1_ACM2017, part2_ICCV2017, part3_ICCV2017, part4_ECCV2018,choi2019hi} aims to match IDs of a person-of-interest across multiple distinct camera views.
These days video-based re-ID \cite{videosup1_CVPR2006, ilids_ECCV2014, videosup3_CVPR2018, videosup4_CVPR2018, videosup5_CVPR2018, videosup6_CVPR2018, videosup7_ICCV2017} has been extensively studied in video surveillance systems for public safety.
Among the re-ID methods, supervised learning approaches lead to substantial  performance improvement.
However, annotating person IDs across multiple cameras entails significantly high labor costs.
This time-consuming labeling work makes re-identification systems hard to be applied in real-world situations since the cost increases dramatically as the number of cameras increases.
Therefore, a line of work \cite{unsup1_ICCV2016, unsup2_CVPR2018, unsup3_CVPR2010, unsup4_ECCV2016, B1_ICCV2017, CAMEL_ICCV2017,  BC2_ECCV2018} focuses on unsupervised re-ID approaches that propose to learn ID-discriminative feature representations without cross-camera person ID labeling.

\begin{figure}[t]
     
     \centering
         \includegraphics[width=0.48\textwidth]{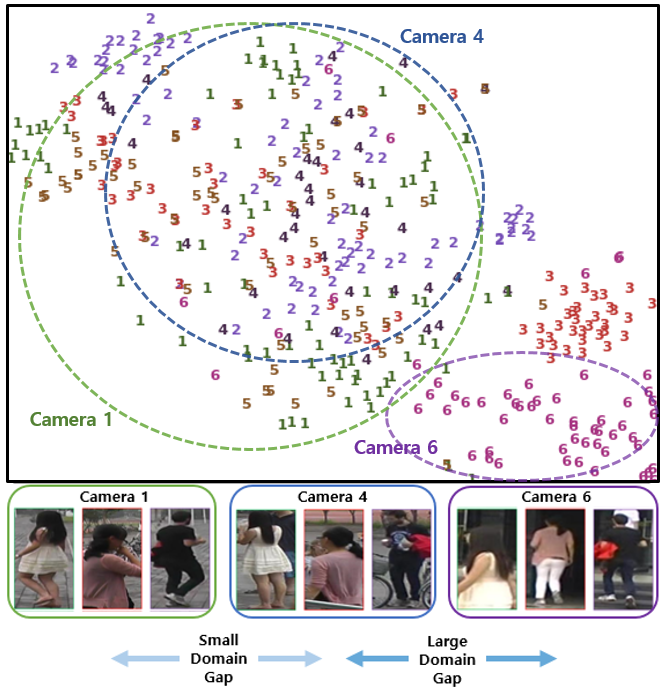}
     \caption{
     We map the features of people extracted from Imagenet pre-trained neural networks to a 2-D space using t-SNE \cite{t-SNE_JMLR2008}.   
     The numbers in different colors denote camera domains.
     It shows that a large domain gap induces the cluster within a camera, which implies that controlling the camera domain gap is important to improve the ID discrimination ability of re-identification systems.
     }
     \label{motivation}
\end{figure}

The major obstacle of unsupervised person re-identification is camera domain discrepancy generated by the difference of viewpoint and background  as shown in Fig. \ref{motivation}.
Recent studies focus on training a model from an additional labeled source dataset and transfer the knowledge to unlabeled target camera domains \cite{PUL_TMM2018, unsup1_ICCV2016, TJ-AIDL_CVPR2018} or iteratively updating a training set using reliable top-$K$ samples across multiple-camera domains \cite{BC1_ICCV2017,B1_ICCV2017, BC2_ECCV2018, C1_ECCV2018, CAMEL_ICCV2017}.
However, performance degradation occurs while transferring knowledge between different domains.
Also, adopting top-$K$ sampling is hard to select reliable samples when the camera domain gap is large. 
Therefore, even though the aforementioned two groups show promising performance in small camera systems \cite{ilids_ECCV2014, prid_SCIA2011}, the nature of imperfect translation and domain locality degrades the performance of large camera systems \cite{mars_ECCV2016, market_ICCV2015}.

The main motivation of this paper is to construct a feature space that considers the relationship of all cameras without information transfer between camera domains.
To this end, we propose a simple yet effective approach, named Parts-Aware camera Domain Alignment Learning (PADAL).
With a carefully designed architecture, we first maximize the distance between different ID features at each camera domain by using a metric learning approach.
While maintaining a feature discrimination ability, we obtain the domain-invariant features by adopting the concept of domain adversarial learning \cite{da1_NIPS2007, da2_JMLR2007}.
Note that our work differs from conventional domain adaptation methods in the point that every camera domain has a different number of IDs. 
This is because not all people move through all cameras in the system.
For example, considering a four-camera system in a real application, it is possible that a person moves through cameras 1, 2, and 4 but does not appear in the camera 3.

To further improve the effect of multi-camera domain alignment, we propose a Part-aware Adaptation Module (PAM).
Most images used in person re-identification are obtained by the detection algorithm \cite{DPM_TPAMI2009} or hand labeling.
Therefore, it is natural that each body part is located in a similar area in the re-ID images. 
For example, \textit{head}, \textit{torso}, and \textit{legs} usually appear on the top, middle, and bottom part of images, respectively.
We leverage this prior information to reduce camera domain discrepancy.
Technically, we make the clusters of the similar spatial features from the embedding feature maps by using the unsupervised clustering algorithm (i.e., K-means).
According to the locations of clustered spatial features, we assign a specialized part-domain discriminator for each cluster.
We find that aligning part features more effectively generates camera-invariant features than aligning global features.

To sum up, our main contributions can be summarized as follows: 
1) We propose a novel learning scheme for unsupervised video person re-identification, named PADAL, which aims to construct the embedding space that represents camera-invariant and ID-discriminative features.
2) To enhance the effectiveness of PADAL, we present a Part-aware Adaptation Module (PAM) to effectively minimize the discrepancy from multiple camera domains by aligning part-level distributions.
3) We conduct extensive experiments on three public video-based person re-ID datasets (PRID-2011 \cite{prid_SCIA2011}, iLIDS-VID \cite{ilids_ECCV2014}, and MARS \cite{mars_ECCV2016}).
Our experimental results show that PADAL is more effective on a large camera setting than other methods, which is an advantage to a real-world application.
The proposed method improves video re-ID performance on the \textit{large-scale} MARS dataset up to {20}\% in rank-1 accuracy.


\section{Related Work}

\subsection{Supervised Person Re-ID}

Following the recent success of deep learning in various fields~\cite{kim2019cnn,kim2020rpm,videosup5_CVPR2018, videosup6_CVPR2018, videosup7_ICCV2017, imgsup_CVPR2018}, 
person re-ID recently has attracted attention due to the importance of video surveillance systems.
In the supervised person re-ID setting \cite{videosup1_CVPR2006, videosup4_CVPR2018, videosup3_CVPR2018, ilids_ECCV2014, videosup5_CVPR2018, videosup6_CVPR2018, videosup7_ICCV2017, imgsup_CVPR2018}, one can use cross-camera label information for training the model which learns camera-invariant features.
To more effectively use the cross-camera labels, \cite{attribute_ECCV2016} proposes the training method that leverages attribute annotations on a person image.
Other studies that use part information \cite{part1_ACM2017, part2_ICCV2017, part3_ICCV2017, part4_ECCV2018} and attention mechanism \cite{attention1_CVPR2018, attention2_CVPR2018} also improve the performance of re-identification.
However, supervised re-identification methods require time-consuming labeling work, which is hard to be applied to large-scale camera surveillance systems.

\subsection{Unsupervised Person Re-ID}

The large amount of effort required for supervised learning leads to the study of unsupervised re-ID \cite{unsup1_ICCV2016, unsup2_CVPR2018, unsup3_CVPR2010, unsup4_ECCV2016, B1_ICCV2017, C1_ECCV2018, CAMEL_ICCV2017, BC1_ICCV2017, BC2_ECCV2018}.
The major obstacle of unsupervised person re-identification is camera domain discrepancy generated by the difference of viewpoints and backgrounds.
The existing unsupervised re-ID approaches for overcoming the discrepancy problem  can be divided into two groups. 
The first group \cite{PUL_TMM2018, unsup1_ICCV2016, TJ-AIDL_CVPR2018} proposes to train a model from an additional labeled source dataset and transfers the knowledge to unlabeled target camera domains.
However, performance degradation occurs while transferring knowledge to different domains and their methods also require an additional labeling cost for the source dataset.
The other group \cite{BC1_ICCV2017,B1_ICCV2017, BC2_ECCV2018, C1_ECCV2018, CAMEL_ICCV2017} suggests to iteratively update a training set using reliable top-K samples across multiple-camera domains.
Unfortunately, a large domain gap would cause the strategy to select different ID samples, which is incorrect.
Even though the aforementioned two groups of work show promising performance in small camera systems (PRID-2011 \cite{prid_SCIA2011} and iLIDS-VID \cite{ilids_ECCV2014}), the nature of imperfect translation and domain locality degrades the performance of large camera systems (MARS \cite{mars_ECCV2016} and Market-1501 \cite{market_ICCV2015}).
Compared to previous work, our method shows a significant performance gain as the number of cameras increases in video surveillance systems, which is appropriate for real-world situations. 
Moreover, we do not use labeled datasets, which is a big advantage in terms of scalability.

\subsection{Domain Adaptation}

Domain adaptation \cite{da1_NIPS2007, da2_JMLR2007} handles domain discrepancy between the labeled source data and the unlabeled target data.
Before the deep learning era, a number of methods \cite{shallow_f1_CVPR2012, shallow_f2_CVPR2013} attempt to decrease domain discrepancy with hand-crafted features.
With the success of deep learning methods, Maximum Mean Discrepancy (MMD) minimization \cite{da_cls1_ICCV2017, da_cls2_JMLR2015} and domain adversarial learning  \cite{adv1_ICML2015, da_seg4_CVPR2018} have achieved impressive performance on the domain adaptation tasks. 
Recently, domain adaptation has been actively studied due to the data scarcity problem in classification \cite{da_cls1_ICCV2017, da_cls2_JMLR2015, da_cls3_ECCV2016}, segmentation \cite{da_seg1_ICCV2017, da_seg2_ICCV2017, da_seg3_CVPR2018, da_seg4_CVPR2018}, and detection \cite{da_det1_CVPR2018, da_det2_CVPR2018}.
In this paper, we apply the concept of domain adversarial learning to unsupervised video re-ID.
Note that our work differs from conventional domain adaptation methods, which assumes the same number of source and target classes.
Compared to these studies, a person randomly appears on each camera so that we do not assume every camera domain has the same number of person IDs.
Also, previous works mainly focus on the classification, segmentation, and detection, which is not a retrieval problem.
Multi-adversarial domain adaptation \cite{md1_AAAI2018} recently proposed to consider the multimode structures underlying the data distribution.
However, they still consider only the source-target domains, which is not applicable in multi-camera domain alignment learning.

\section{Methodology}

Our goal is to train a network to extract ID features which are invariant to camera domains while preserving an ID discrimination ability.
The proposed approach consists of three main parts as follows:
1) For accurate ID classification, the network extracts features that have a large distance between different IDs. 
To this end, we adopt a metric learning approach.
2) Multi-camera domain invariant feature learning generates camera-invariant ID features. 
3) Moreover, we observe that considering person parts during multi-domain alignment induces more effective domain alignment performance.

After training, the network extracts features from given query/gallery tracklets (i.e., a sequence of person images), and the features are compared by the Euclidean distance for ranking.
Our approach can be applied not only to person re-identification but also to various multi-domain tasks with prior information of an input image.

\begin{figure}[t]
     
     \centering
         \includegraphics[width=0.48\textwidth]{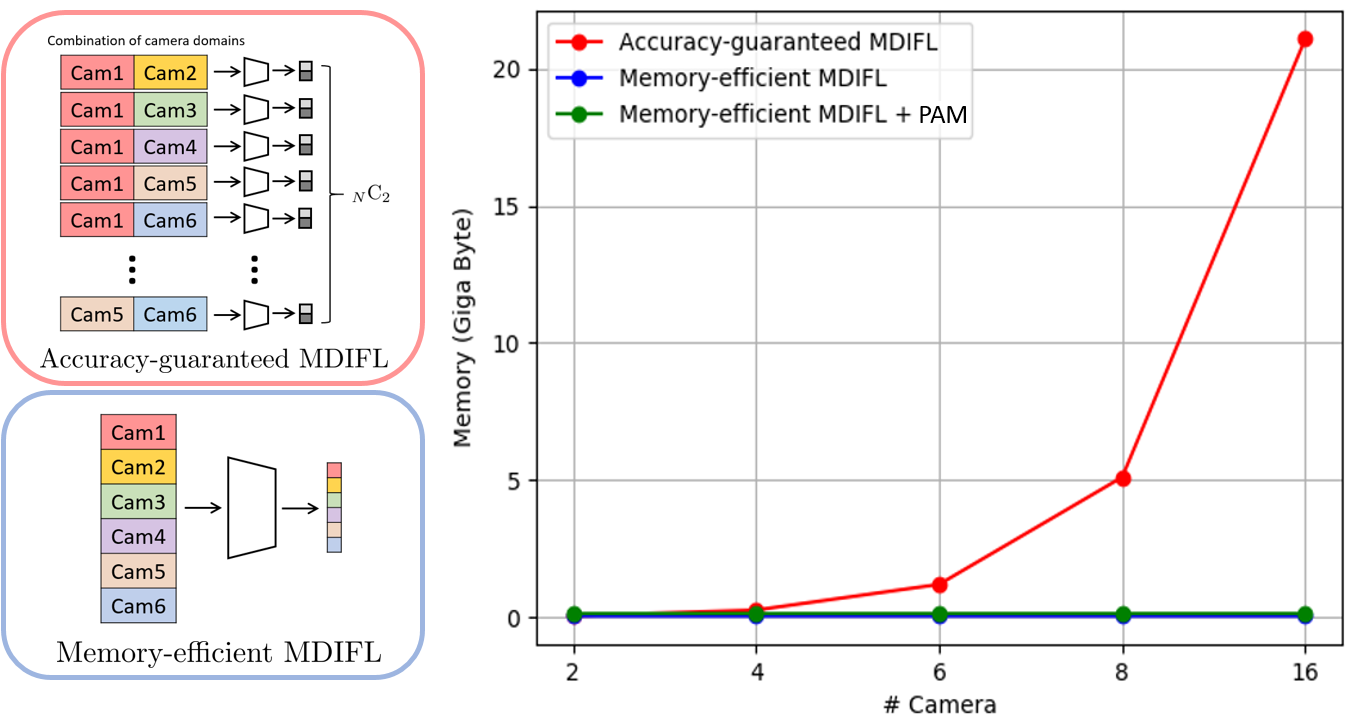}
    \caption{Illustration of two versions of aligning multi-camera domains: Accuracy-guaranteed MDIFL and Memory-efficient MDIFL. 
    Even though Accuracy-guaranteed MDIFL shows better performance than Memory-efficient MDIFL, the memory requirement increases in $O(n^2)$, where $n$ is the number of cameras.
    Adding our PAM to Memory-efficient MDIFL fills the performance gap without the significant memory consumption.
    }
     \label{MDISC}
\end{figure}

\subsection{Problem Formulation}
Video person re-identification (re-ID) \cite{videosup1_CVPR2006, ilids_ECCV2014} is a task of searching the given probe person in the gallery video sets.
Since each camera has different camera parameters and viewpoints, we set each camera to a different domain.
We denote ${C} = \{1, \cdots, N_{c}\}$ as camera domain set, each domain contains a different tracklet set ${T_{i}} = \{1, \cdots, N_{T_{i}}\}$, $i \in C$.
Here, $ N_{c}$ and $N_{T_{i}}$ are the number of cameras and the number of tracklets in camera $i$, respectively. 
Let $D = \{({\bm{I}}_{(i, j)}, Y_{(i, j)})_{j=1}^{N_{T_{i}}}\}_{i=1}^{ N_{c}}$ represent the tracklet-label pairs for network training, where $\bm{I}_{(i, j)}$ is the \textit{j}-th tracklet from the \textit{i}-th camera  and $Y_{(i, j)}$ is the ID label  in the form of one-hot-vectors.
Usually, supervised person re-identification methods exploit ID information between cameras to find the camera-invariant feature of tracklets.
Compared to these methods, our work aims to train the network without ID information between cameras.

\subsection{Improving an ID Discrimination Ability}
\label{section3.2}

Even though there is no ID information across multiple cameras, the network can learn ID-discriminative feature representation.
We adopt a metric learning approach for each camera domain, which shares a common feature space.
The information that we require for metric learning is whether each tracklet is from the same or different person in one camera.
Therefore, in each camera domain, we randomly assign the different labels to each tracklet. 
For example, considering that the tracklet label $1$ in camera $1$ and $2$, $Y_{(1, 1)}$ and $Y_{(2, 1)}$ might not represent the same person.

Technically, we first embed the input space $S$ into the feature space $F$ regardless of its camera domain by using a common feature extractor network $f: S \rightarrow F$.
Then embedding features in the space $F$ are projected onto other feature spaces $G_{i}$ according to the camera domain $i \in C$.
Here, we use the fully-connected layer $g_i: F \rightarrow G_{i}$, i.e., matrix multiplication, for linear projection.
Since pseudo-labels are assigned to tracklets in each camera domain, we can calculate the cross-entropy loss of samples in the space $G_{i}$.
We minimize an ID-discriminative loss $\mathcal{L}_{id}$ which is defined as follows:

\begin{equation}
    \label{Lid}
       \mathcal{L}_{id}(\bm{I}_{(i,j)}) = 
       -\frac{1}{ N_{T_{i}}}\sum_{p\in{T_i}} Y_{(i,j)}^p log(g_i(f(\bm{I}_{(i, j)})))^p.
\end{equation}

In our work, the order of pseudo-labeling does not affect on the performance of re-identification since we minimize the cross entropy function regarding samples in the dedicated space $G_{i}$.
Note that we use features in the space $F$ during the test stage.
Without using cross-camera person ID information,
the trained feature extractor $f$ has an ID discrimination ability in the space $F$.
Nevertheless, misalignment of camera domains still degrades re-identification performance (see Fig. \ref{result_tsne}).

\subsection{Learning to Align Camera Domains}
\label{section 3.3}

The key contribution of this paper is proposing effective Multi-camera Domain Invariant Feature Learning (MDIFL),
which enforces the feature extractor $f(\cdot)$ to generate the similar ID features across all camera in the space $F$. 
We apply the concept of domain adversarial learning to implement MDIFL by considering the two characteristics of person re-ID.
First, in video surveillance systems, there are usually more than two camera domains, which is hard to apply the conventional source to target domain adaptation approach.
Therefore, when designing a domain discriminator, we should consider the scalability of the re-identification system.
Second, we can utilize the prior information of a structured person image obtained by the detection algorithm \cite{DPM_TPAMI2009} or hand labeling.
We first introduce two versions of MDIFL and then explain our Part-aware Adaptation Module (PAM).

\textbf{Accuracy-guaranteed MDIFL.}
\hspace{2mm}
We design the camera domain discriminator $h_{(u,v)}(\cdot)$  between every pair of domains, where $u \neq v$ and $u,v  \in  \{1, \cdots, N_{c} \}$.
Note that both $h_{(u,v)}(\cdot)$ and $h_{(v,u)}(\cdot)$
represent the same camera domain discriminator.
For instance, considering a re-identification system with three cameras, the network requires three discriminators, i.e.,  $h_{(1,2)}(\cdot)$, $h_{(1,3)}(\cdot)$, and $h_{(2,3)}(\cdot)$. 
We formulate an adversarial loss function for an input tracklet $\bm{I}_{(i,j)}$ from camera domain $i$ as follows: 

\begin{equation}
    \label{La}
    \mathcal{L}_{a}(\bm{I}_{(i,j)}) = -  \sum_{k \neq i}  log(h_{(i,k)}(f(\bm{I}_{(i,j)}))).
\end{equation}

Accuracy-guaranteed MDIFL shows higher performance than a memory-efficient version since it uses more resources to distinguish camera domains.
However, cross-domain discriminators for $n$ cameras have a memory complexity $O(n^2)$, which limits the number of cameras for video surveillance systems.
For example, as shown in Fig. \ref{MDISC}, the 6 camera system requires 15 discriminators ($\comb{6}{2}$) for training domain invariant features.

\textbf{Memory-efficient MDIFL.}
To effectively implement MDIFL, we design a camera discriminator to take the ID features of multiple domains with shared parameters.
Therefore, a number of the cross-domain discriminators can be compressed into one multi-domain discriminator. 
The discriminator produces the output as the probability of each domain.

We know the camera domain of each input tracklet $\bm{I}_{(i,j)}$ from camera $i$ so that we can define the camera domain label $D_{i} \in \mathbb{R}^{N_{c}}$ in the form of one-hot-vectors.
Therefore, the objective function for MDIFL with a single discriminator is formulated as follows:

\begin{equation}
    \label{Lm}
       \mathcal{L}_{m}(\bm{I}_{(i,j)}) = 
       -\sum_{c\in C} D_{i}^c log(h(f(\bm{I}_{(i, j)})))^c.
\end{equation}

We observe that designing a multi-domain discriminator requires a performance-memory trade-off by our experiments.
The proposed $\mathcal{L}_{m}(\bm{I}_{(i,j)})$ leads $f(\cdot)$ to generate inter-camera invariant ID features.

\begin{figure}[t]
     \centering
         \includegraphics[width=0.46\textwidth]{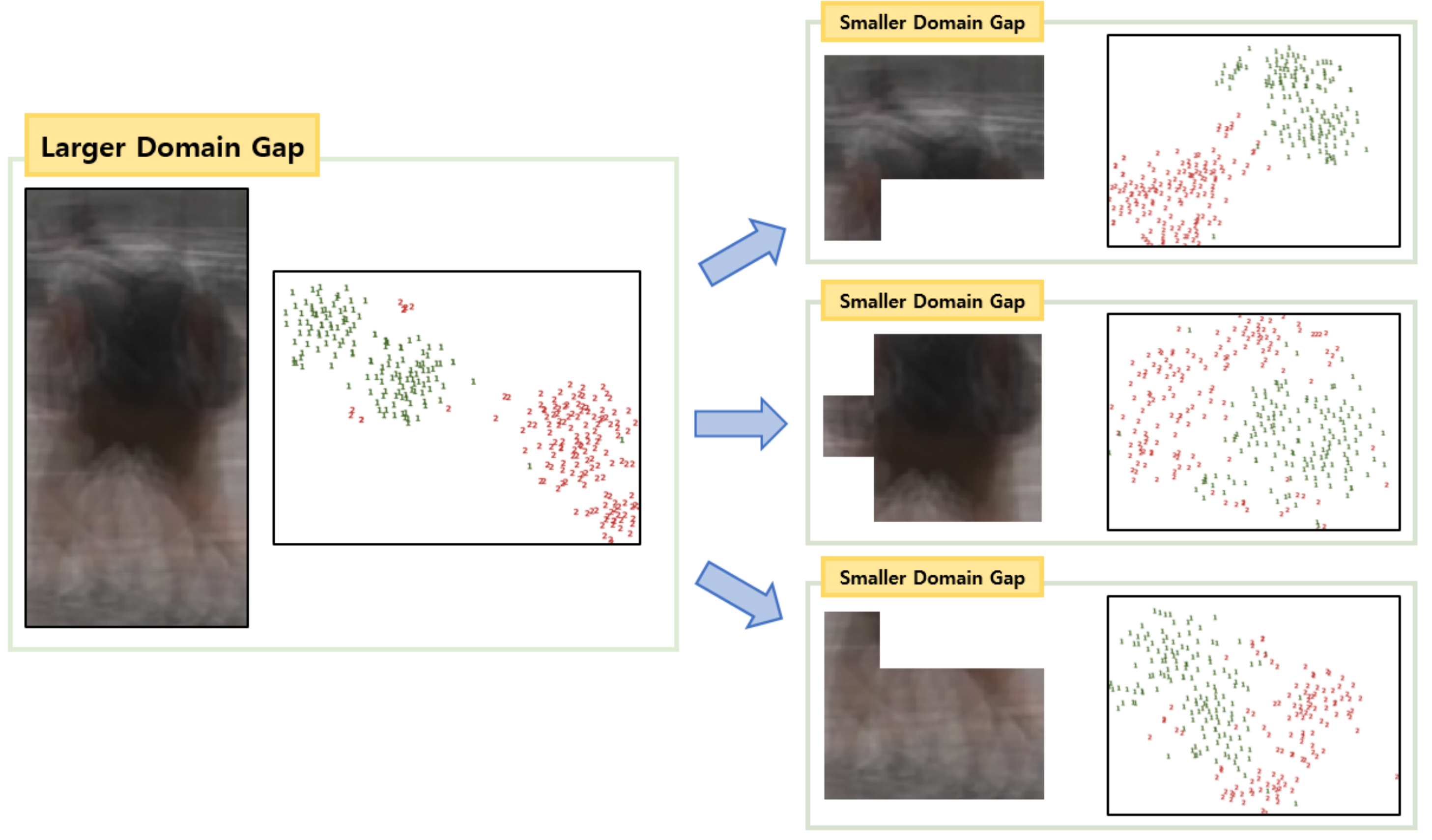}
    \caption{
    We observe that spatially-clustered parts have a smaller domain gap compared to global information. 
    We map the global and parts features to a 2-D space by using t-SNE \cite{t-SNE_JMLR2008}.   
    The element number displayed on t-SNE denotes camera domain.
    Since we use the video sequence for re-identification, we can get reliable person features by averaging features along the time axis. 
    }
     \label{PPA_obs}
\end{figure}

\begin{figure*}[t]
     \centering
         \includegraphics[width=0.95\textwidth]{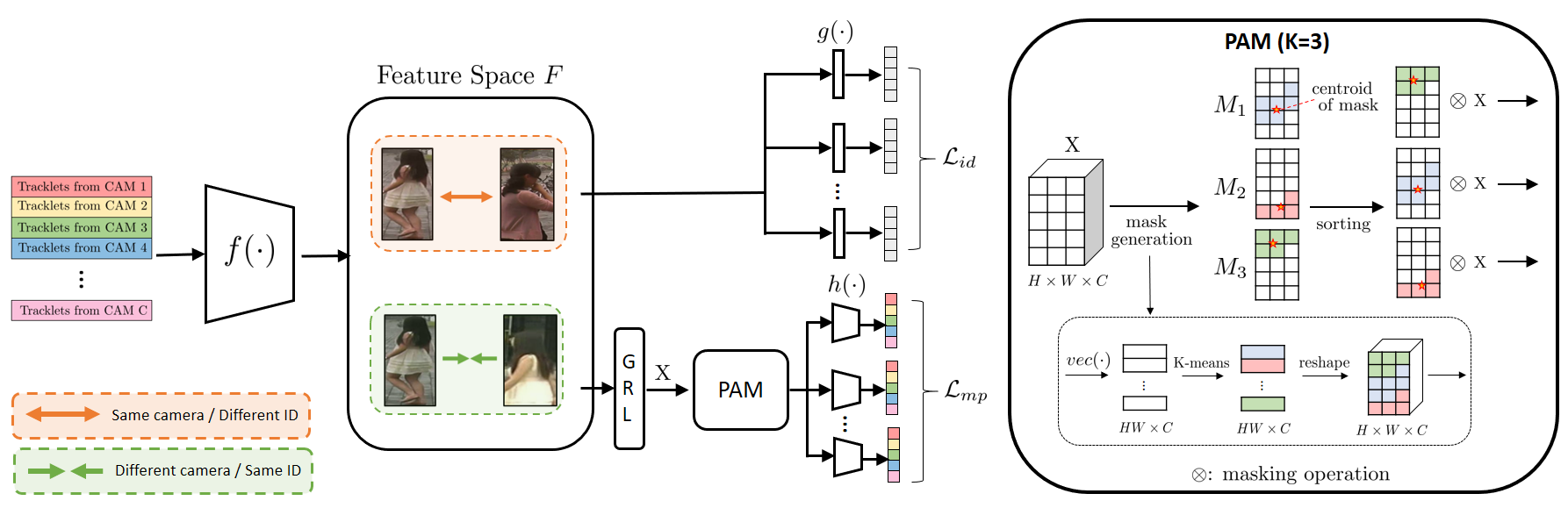}
\vspace{1mm}
 \caption{Illustration of the network architecture.
 We first embed tracklets into the feature space $F$.
 In this space, we increase the distance between the different ID features in the same camera domain (orange) by minimizing the loss $\mathcal{L}_{id}$.
 At the same time, we reduce the distance between the same ID features across different camera domains (green) by minimizing the loss $\mathcal{L}_{mp}$.
 Our PAM clusters similar spatial features to generate masks and then sort the order of masks. Finally, the feature distribution of each part is aligned by a specialized camera domain discriminator.
 }
     \label{Main_pipeline}
\end{figure*}

\subsection{Part-aware Adaptation Module}
\label{section 3.4}
Previous domain adaptation methods consider image-level alignment (i.e., using the average-pooled feature) for image classification tasks and pixel-level alignment for semantic segmentation tasks.
However, aligning image or pixel-level features might not be the optimal choice for feature alignment regarding structured images such as person images.
This is because each body part has a different feature distribution with respect to multiple camera domains.
Moreover, we observe that spatially-clustered parts have a smaller domain gap between different cameras compared to the global information of the image as shown in Fig. \ref{PPA_obs}.
Therefore, we suggest to cluster similar spatial features and independently align these features. 
Note that we explain the PAM based on the assumption that we use memory-efficient MDIFL.

To cluster similar spatial features into $K$ clusters  from embedding feature $\mathbf{X} \in \mathbb{R}^{H \times W \times C}$, we use the K-means clustering algorithm.
Based on clustered locations, we generate the person part mask $M_k \in \mathbb{R}^{H \times W }$, where $k  \in  \{1, \cdots, K \}$.
Since using the K-means algorithm does not consider label order, we design a sorting algorithm.
Interestingly, we observe that clustered features construct a horizontal line as shown in Fig. \ref{kpartshow}.
Using this observation, we sort the masks based on the y coordinate (i.e., height) of the centroid of the mask.
Then specialized part-domain discriminators $h^{k}(\cdot)$ take masked feature maps as an input.
Each specialized part-domain discriminator consists of fully-convolutional layers so that it produces the pixel-wise probability regarding the camera domain.
We can formulate the discriminator loss as follows:
\begin{small}
\begin{equation}
       \mathcal{L}_{mp}(\bm{I}_{(i,j)}) = 
    -\hspace{-3mm}\sum_{k,h,w,c}\hspace{-3mm}
       M_{k}^{(h,w)} \hspace{-1mm} D_{i}^{(h,w,c)} log(h^{k}(M_{k} \otimes \mathbf{X}_{(i,j)}))^{(h,w,c)},
      \label{Lmp}
\end{equation}
\end{small}
where $\otimes$ represents a masking operation.

We show that setting K to 3 achieves the best performance in our experiment section.
This is related the fact that \textit{head}, \textit{torso}, and \textit{legs} usually appear on the top, middle, and bottom part of an image, respectively.
We illustrate our PAM in Fig. \ref{Main_pipeline}.
In the same way, PAM can be applied to accuracy-guaranteed MDIFL.

\begin{table*}[t]
\centering
\begin{tabular}{l|cccc|cccc|ccccc}
 \Xhline{3\arrayrulewidth}

Methods    & \multicolumn{4}{c|}{PRID-2011 ($\# CAM = 2$)}                                                                              & \multicolumn{4}{c|}{iLIDS-VID ($\#CAM = 2$)}                                                                             & \multicolumn{5}{c}{MARS ($\#CAM = 6$)}                              \\ \hline
Rank        & 1                        & 5                        & 10                       & 20                        & 1                        & 5                        & 10                       & 20                        & 1    & 5    & 10   & \multicolumn{1}{c|}{20}   & mAP  \\ \hline
Salience \cite{salience_CVPR2013}   & \multicolumn{1}{l}{25.8} & \multicolumn{1}{l}{43.6} & \multicolumn{1}{l}{52.6} & \multicolumn{1}{l|}{62.0} & \multicolumn{1}{l}{10.2} & \multicolumn{1}{l}{24.8} & \multicolumn{1}{l}{35.5} & \multicolumn{1}{l|}{52.9} & -    & -    & -    & \multicolumn{1}{c|}{-}    & -    \\
STFV3D  \cite{compare_add1_ICCV2015}      & 27.0                     & 54.0                     & 66.3                     & 80.9                      & 19.1                     & 38.8                     & 51.7                     & 70.7                      & - & - & - & \multicolumn{1}{c|}{-} & -  \\
FV3D  \cite{compare_add1_ICCV2015}      & 38.7                     & 71.0 & 80.6                     & 90.3                      & 25.3                     & 54.0                     & \textbf{\textcolor{blue}{68.3}}                      & \textbf{\textcolor{red}{87.3}}  & - & - & - & \multicolumn{1}{c|}{-} & -  \\
DVDL  \cite{compare_add2_ICCV2015}      & 40.6                     & 69.7                     & 77.8                     & 85.6                      & 25.9                     & 48.2                     & 57.3                     & 68.9                       & - & - & - & \multicolumn{1}{c|}{-} & -  \\
GRDL  \cite{unsup4_ECCV2016}      & 41.6                     & 76.4                     & 84.6                     & 89.9                      & 21.7                     & 42.9                     & 56.2                     & 71.6                      & 19.3 & 33.2 & 41.6 & \multicolumn{1}{c|}{46.5} & 9.6  \\
SMP  \cite{BC1_ICCV2017}         & \textbf{\textcolor{red}{80.8}}                   & \textbf{\textcolor{red}{96.0}}                    & \textbf{\textcolor{red}{98.3}}                   & \textbf{\textcolor{red}{99.3}}                    & 33.2                     & 55.7                     & 64.4                     & 72.5                      & 23.9 & 35.8 & -    & \multicolumn{1}{c|}{44.9} & 10.5 \\
DGM+IDE  \cite{B1_ICCV2017}   & 56.4                     & 81.3                     & 88.0                       & 89.9                      & \textbf{\textcolor{red}{36.2}}                      & \textbf{\textcolor{red}{62.8}}                      & \textbf{\textcolor{red}{73.6}}                      & {82.7}                  & 36.8 & 54.0 & 61.6 & \multicolumn{1}{c|}{68.5} & 21.3 \\
RACE   \cite{BC2_ECCV2018}      & 50.6                     & 79.4                     & 84.8                     & 91.8                      & 19.3                     & 39.3                     & 53.3                     & 68.7                      & 43.2 & 57.1 & 62.1 & \multicolumn{1}{c|}{67.6} & 24.5 \\
TAUDL \cite{C1_ECCV2018}       & 49.4                     & 78.7                     & -                        & \textbf{\textcolor{blue}{98.9}}                      & 26.7                     & 51.3                     & -                        & 82.0                      & 43.8 & 59.9 & -    & \multicolumn{1}{c|}{72.8} & 29.1 \\ \hline
PADAL${_{mp}}$ (Ours)
&            \textbf{\textcolor{blue}{67.3}}          &      \textbf{\textcolor{blue}{87.8}}                    &              \textbf{\textcolor{blue}{93.7}}            &              97.1             & {\textbf{\textcolor{blue}{34.4}}}                     & {\textbf{\textcolor{blue}{55.4}}}                     & 67.1                   & {\textbf{\textcolor{blue}{83.2}}}                    
& \textbf{\textcolor{blue}{61.7}} & \textbf{\textcolor{blue}{77.7}} & \textbf{\textcolor{blue}{83.4}} & \multicolumn{1}{c|}{\textbf{\textcolor{blue}{87.9}}} & \textbf{\textcolor{blue}{48.3}}\\  
PADAL${_{a}}$  (Ours) &         -         &      -                  &              -         &              -            & -                   & -                    &-                 & -                            & \textbf{\textcolor{red}{63.3}} & \textbf{\textcolor{red}{80.8}} & \textbf{\textcolor{red}{85.7}} & \multicolumn{1}{c|}{\textbf{\textcolor{red}{89.8}}} & \textbf{\textcolor{red}{51.7}}\\  \hline

Supervised  \cite{compare_sup_CVPR2016}       & 85.2                    & 97.1                     & 98.9                     & 99.6                      & 60.2                     & 84.7                     & 91.7                     & 95.2                      & 71.2 & 85.7 & 91.8    & \multicolumn{1}{c|}{94.3} & - \\

 \Xhline{3\arrayrulewidth}
\end{tabular}
 \hspace{20mm}
  \caption{Performance evaluation on the three unsupervised video re-ID datasets: PRID-2011 (two cameras), iLIDS-VID (two cameras), and MARS (six cameras).
    Here, we denote the accuracy-guaranteed and the memory-efficient version with PAM as PADAL$_{a}$ and PADAL$_{mp}$, respectively.
    $1^{st}$ and $2^{nd}$ best results are indicated in \textbf{\textcolor{red}{red}} and \textbf{\textcolor{blue}{blue}} colors respectively.
    }
\label{sota_comp}
\end{table*}

\subsection{Network Architecture and Training}

\textbf{Baseline Network.}
We illustrate our network in Fig. \ref{Main_pipeline}.
For a fair comparison with previous methods, we adopt the ResNet-50 \cite{Resnet_CVPR2016} model pre-trained on ImageNet as our ID feature extractor $f(\cdot | \theta_f)$.
Every intra-camera ID classifier $g_{i}(\cdot| \theta_{g_i})$ consists of a single fully-connected layer followed by a softmax layer.
The GRL \cite{adv1_ICML2015} layer changes the gradients sign of feature $\mathbf{X}$ to negative when back-propagation is performed.

\textbf{Camera Domain Discriminator.}
To consider the part feature distributions, we adopt fully-convolutional layers, which maintain spatial information.
Specifically,
the discriminator network consists of 3 convolutional layers with a kernel $3 \times 3$, a stride of 1, and a zero-padding of 1.
The channel size of each convolution layer is  $\left \{2048, 512, 256, N_{c}  \right \}$, where $N_{c}$ is the number of camera domains.
We use Leaky ReLU parameterized by 0.2 for the activation function.

\textbf{Network Training.}
We build our work based on domain adversarial learning, which consists of two-phase learning.
In the first phase, the feature extractor $f(\cdot|\theta_f)$ learns to generate embedding features that confuse a domain discriminator.
In the second phase, the domain discriminator $h(\cdot|\theta_h)$ is trained to distinguish between source and target domains.

Specifically, we train the parameters $\theta_f$ of the feature extractor  to maximize the discriminator loss $\mathcal{L}_{mp}$ in eq. \ref{Lmp} (or we can use variants of the MDIFL loss: $\mathcal{L}_{a}$, $\mathcal{L}_{m}$, and $\mathcal{L}_{ap}$).
Also, the parameters $\theta_{g}$ of fully-connected layers for ID-discriminative learning are trained to minimize  $\mathcal{L}_{id}$ (eq. \ref{Lid}).
On the other hand, the parameters $\theta_{h}$ of the camera domain discriminators are learned by minimizing the loss $\mathcal{L}_{mp}$.
The training objective for unsupervised person re-identification can be formulated as follows:
\begin{equation}
    \label{all_loss}
     \mathcal{L}(\bm{I}_{(i,j)}) = \mathcal{L}_{id}(\bm{I}_{(i,j)}) - \lambda \mathcal{L}_{mp}(\bm{I}_{(i,j)}),
\end{equation}
where $\lambda$ is a hyperparameter for balancing the two loss functions.
Based on eq. \ref{all_loss}, we find the parameters $\hat{\theta}_f, \hat{\theta}_g, \hat{\theta}_h$ at a saddle point:
\begin{equation}
    \label{minmax}
    (\hat{\theta}_f, \hat{\theta}_g) = \argmin_{\theta_f, \theta_g}  \mathcal{L}(\bm{I}_{(i,j)}).
\end{equation}
\begin{equation}
    \label{minmax}
    \hat{\theta}_h =  \argmax_{\theta_h}  \mathcal{L}(\bm{I}_{(i,j)}).
\end{equation}
In our work, we use the gradient reversal layer (GRL) \cite{adv1_ICML2015} for implementation.
During the test stage, the network embeds query/gallery tracklets to the feature space $F$, and the features are compared by the Euclidean distance for ranking.

\begin{table}[t]
\centering

\begin{tabular}{l|c|cccc}
\Xhline{3\arrayrulewidth}
Loss        & $K$     & Rank-1   & Rank-5  & Rank-20 & mAP  \\ \hline
$\mathcal{L}_{id}$      & -        & 42.9 & 63.6& 78.8 &30.6 \\
$\mathcal{L}_{id}$ + $\mathcal{L}_{a}$  & -             & 63.3& 80.8& 89.8& 51.7 \\
$\mathcal{L}_{id}$ + $\mathcal{L}_{m}$   & -  & 52.9& 69.6& 83.5& 38.5 \\\hline
\multirow{5}{*}{$\mathcal{L}_{id}$ + $\mathcal{L}_{mp}$}   & 1   & 55.0 & 72.7&  84.3&42.0 \\
                                & 2   & 61.5& 77.4& 88.0& 48.0 \\
                                & 3   & 61.7 & 77.7& 87.9& 48.3 \\
                                & 4   & 59.8 & 77.8& 87.9& 47.1 \\
                                & 5   & 57.6 & 75.3& 87.3& 45.3 \\
 \Xhline{3\arrayrulewidth}
\end{tabular}
\vspace{1ex}
\caption{Ablation study on the MARS dataset. We present the performance of the network with respect to the loss functions.}
 \label{ablation}
\end{table}
\vspace{-2mm}


\section{Experiments}

\textbf{Implementation Details.}
To train the network, we utilized the Adam optimizer \cite{Adam} with a weight decay $5 \times 10^{-4}$.
We used the learning rate of 0.00035 and  decayed the learning rate by 0.1 every 200 training steps.
Note that all training images were resized to $224 \times 112$.
The whole pipeline was implemented by using the Pytorch framework \cite{paszke2017automatic} on two NVIDIA Titan Xp GPUs.

\textbf{Datasets.}
To show the generality of our method, we report the performance on three public video re-ID datasets: PRID-2011 \cite{prid_SCIA2011}, iLIDS-VID \cite{ilids_ECCV2014}, and \textit{large-scale} MARS \cite{mars_ECCV2016}.
People in the PRID-2011 dataset are captured by two disjoint cameras with a large domain gap.
It contains 385 and 749 people video tracks in camera A and camera B, respectively. 
Following the previous protocol, 178 people (i.e., 89 people for each training and testing respectively) with no less than 27 frames were used for evaluation. 
The iLIDS-VID dataset is collected from two non-overlapping cameras located in an airport hall. Three hundred people video tracks are sampled in each camera (i.e., 150 people for each training and testing respectively).
MARS is a large-scale dataset with non-overlapping six cameras.
It contains 1,261 person IDs, 625 for training and 636 for testing.
Since MARS has multiple tracklets per ID in a camera, we assign the same pseudo-ID for multiple tracklets regarding one person.

\textbf{Evaluation Metric.}
We use CMC scores for evaluating all datasets.
Additionally, we compute mean Average Precision (mAP) for the 6-camera MARS dataset.

\begin{figure*}[t]
    
     \centering
         \includegraphics[width=0.9\textwidth]{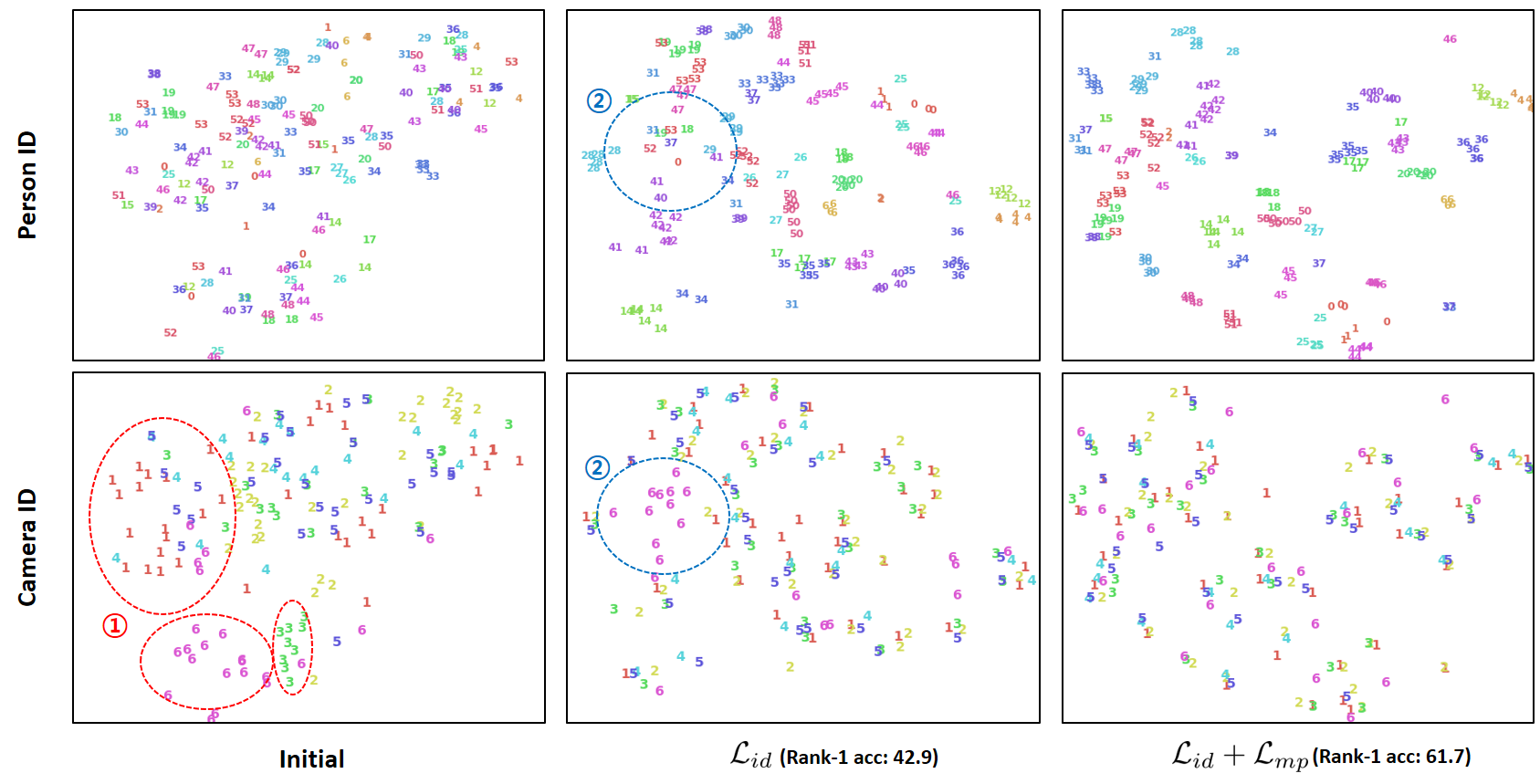}
    \caption{Visualization of the feature space $F$ in terms of person ID (top row) and camera ID (bottom row). Here, we use the samples from the MARS test set.
    In the initial phase, we observe that samples in the same camera domain are located in the similar regions  (\textcolor{red}{\textcircled{\raisebox{-0.9pt}{1}} red circles}).
    With ID-discriminative feature learning (Section \ref{section3.2}), the network can learn ID-discriminative feature representations. 
    However, the misalignment of camera domains is still observed in the feature space (\textcolor{blue}{\textcircled{\raisebox{-0.9pt}{2}} blue circles}).
    To overcome the problem, we propose MDIFL so that we can align the multiple camera domains.  
    }
     \label{result_tsne}
\end{figure*}

\subsection{Comparison with State-of-the-art Methods}
We compare PADAL$_{a}$ (accuracy-guaranteed version), PADAL$_{m}$ (memory-efficient version), and PADAL$_{mp}$ (memory-efficient version + PAM) with state-of-the-art methods, including Salience \cite{salience_CVPR2013}, STFV3D \cite{compare_add1_ICCV2015}, FV3D \cite{compare_add1_ICCV2015}, DVDL \cite{compare_add2_ICCV2015}, GRDL \cite{unsup4_ECCV2016}, SMP \cite{BC1_ICCV2017}, DGM \cite{B1_ICCV2017}, RACE \cite{BC2_ECCV2018}, TAUDL \cite{C1_ECCV2018} on PRID-2011 \cite{prid_SCIA2011}, iLIDS-VID \cite{ilids_ECCV2014}, and \textit{large-scale} MARS \cite{mars_ECCV2016} datasets in Table \ref{sota_comp}.
The results demonstrate that:
(1) For small camera datasets (PRID-2011 and iLIDS-VID), our approach achieves comparable performance with state-of-the-art methods.
SMP \cite{BC1_ICCV2017} shows the better performance on the PRID-2011 dataset than ours, however, it shows the lower performance on the MARS dataset.
This is because the top-K sampling strategy selects reliable neighbors when a small camera domain gap exists, but not in a large camera domain gap.
(2) For the multiple-camera dataset (MARS), our method achieves state-of-the-art performance with a large margin.
Specifically, our PADAL${_{a}}$ exceeds the recent methods (e.g., TAUDL \cite{C1_ECCV2018} and RACE \cite{BC2_ECCV2018}) 
on the MARS dataset by {19.5}\% (63.3-43.8) in rank-1 accuracy and  {22.6}\% (51.7-29.1) in mAP.
Also, our memory-efficient version PADAL${_{mp}}$ outperforms these methods by {17.9}\% (61.7-43.8) in rank-1 accuracy and  {19.2}\% (48.3-29.1) in mAP.
Recall that one can choose between two versions, PADAL${_{a}}$ and PADAL${_{mp}}$, depending on the size of the camera network system.
In Fig. \ref{MDISC}, we show that PADAL${_{mp}}$ is easy-to-expand to a video surveillance system with a large number of cameras.
(3) The other methods are mainly based on the comparison of the feature distance between samples from different camera domains.
These methods show promising performance in a small camera system but not in a large camera network due to the uncertainty from computing distance between different camera domains.
Our method addresses this problem by adopting a direct domain alignment approach with adversarial learning, which is applicable to large camera systems.

\begin{table}[t]
\centering
\begin{tabular}{l|cccc|c}
\Xhline{3\arrayrulewidth}
Methods         & R1   & R5   & R10  & R20  & mAP  \\ \hline
Baseline  & 42.9 & 63.6 & 71.3 & 78.8 & 30.6 \\
2-domains $\times$ 3 & 45.4 & 66.6 & 73.5 & 79.4 & 33.3 \\
3-domains $\times$ 2 & 51.5 & 70.5 & 77.6 & 83.3 & 38.1 \\
6-domains  $\times$ 1  & 55.0 & 72.7 & 79.1 & 84.3 & 42.0 \\
\Xhline{3\arrayrulewidth}
\end{tabular}
 \hspace{20mm}
\caption{ The performance of partial-domain alignment on the MARS dataset. 
We use PADAL$_{mp}$ with $K=1$ for the experiment. 
``$m$-domains $\times$ $n$" denotes $n$ multi-domain discriminators, which of each domain classifier aligns $m$ camera domains.
}
\label{DomainN}
\end{table}

\begin{figure}[t]
    
     \centering
         \includegraphics[width=0.45\textwidth]{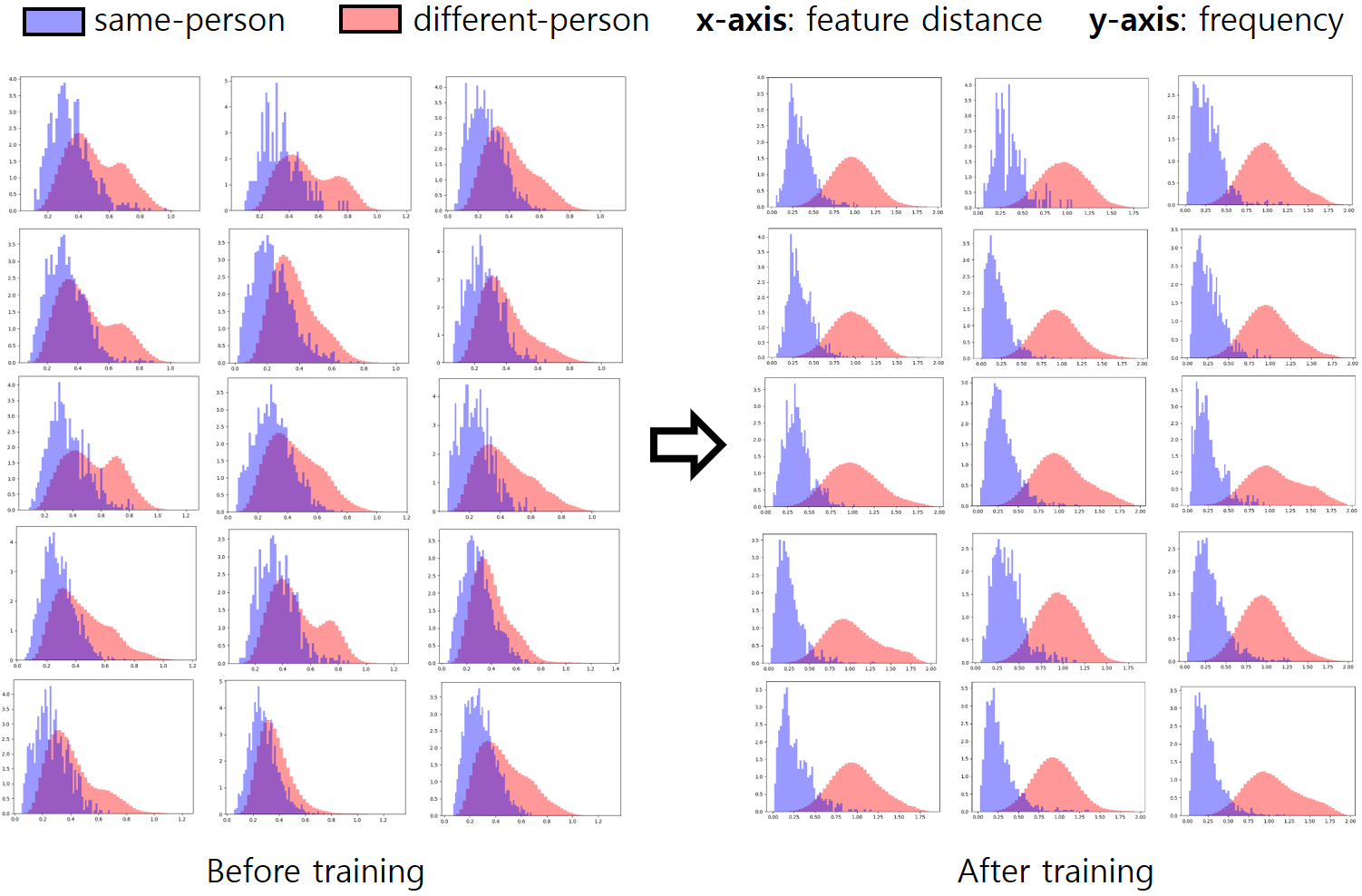}
    \caption{Histogram of the feature distance. We visualize same-person and different-person feature distance obtained from 15 combinations of camera domains in the MARS dataset. 
    }
     \label{histogram}
\end{figure}

\subsection{Analysis of the Proposed Method} \label{analysis_section}

\textbf{Ablation Study.}
In Table \ref{ablation}, we show the effect of different configurations:
PADAL$_{a}$ ($\mathcal{L}_{id}+\mathcal{L}_{a}$), PADAL$_{m}$ ($\mathcal{L}_{id}+\mathcal{L}_{m}$), and PADAL$_{mp}$ ($\mathcal{L}_{id}+\mathcal{L}_{mp}$).
Clearly, PADAL$_{a}$, PADAL$_{m}$, and PADAL$_{mp}$ achieve the higher performance than Baseline ($\mathcal{L}_{id}$), which demonstrates that adversarial learning for aligning multi-camera domain gaps improves the discrimination ability of re-identification systems. 
Moreover, we observe that the accuracy-guaranteed version, which exploits multiple cross-domain discriminators, shows better performance than the memory-efficient version (52.9 $\%$ vs.  63.3 $\%$).
The performance of PADAL$_{mp}$ varies according to the value of the cluster number $K$.

\begin{figure}[t]
    
     \centering
         \includegraphics[width=0.48\textwidth]{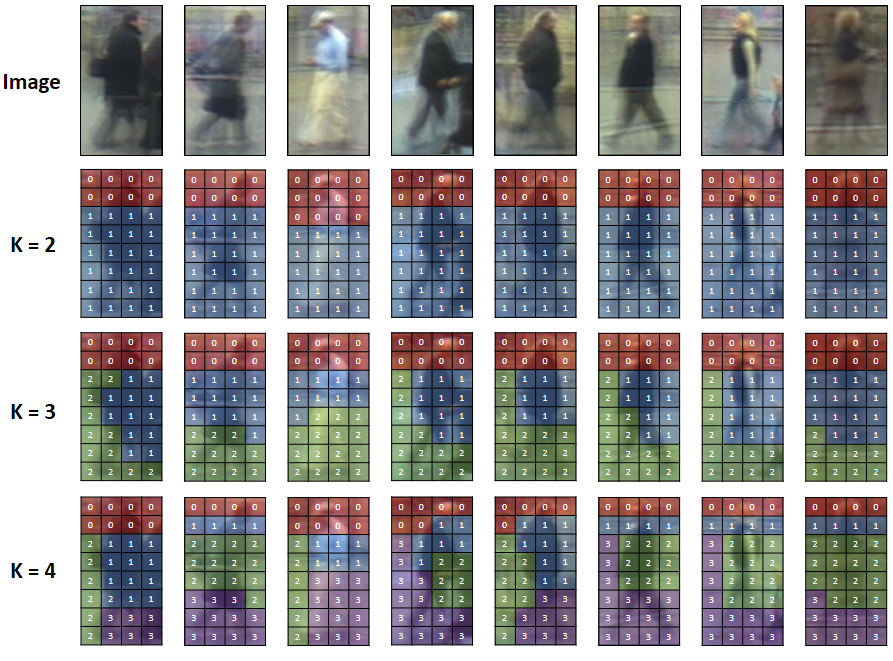}
    \caption{Visualization of the clustered parts by our PAM.
    Different colors and numbers represent different part labels.
    }
     \label{kpartshow}
\end{figure}

\begin{figure}[t]
    
     \centering
         \includegraphics[width=0.48\textwidth]{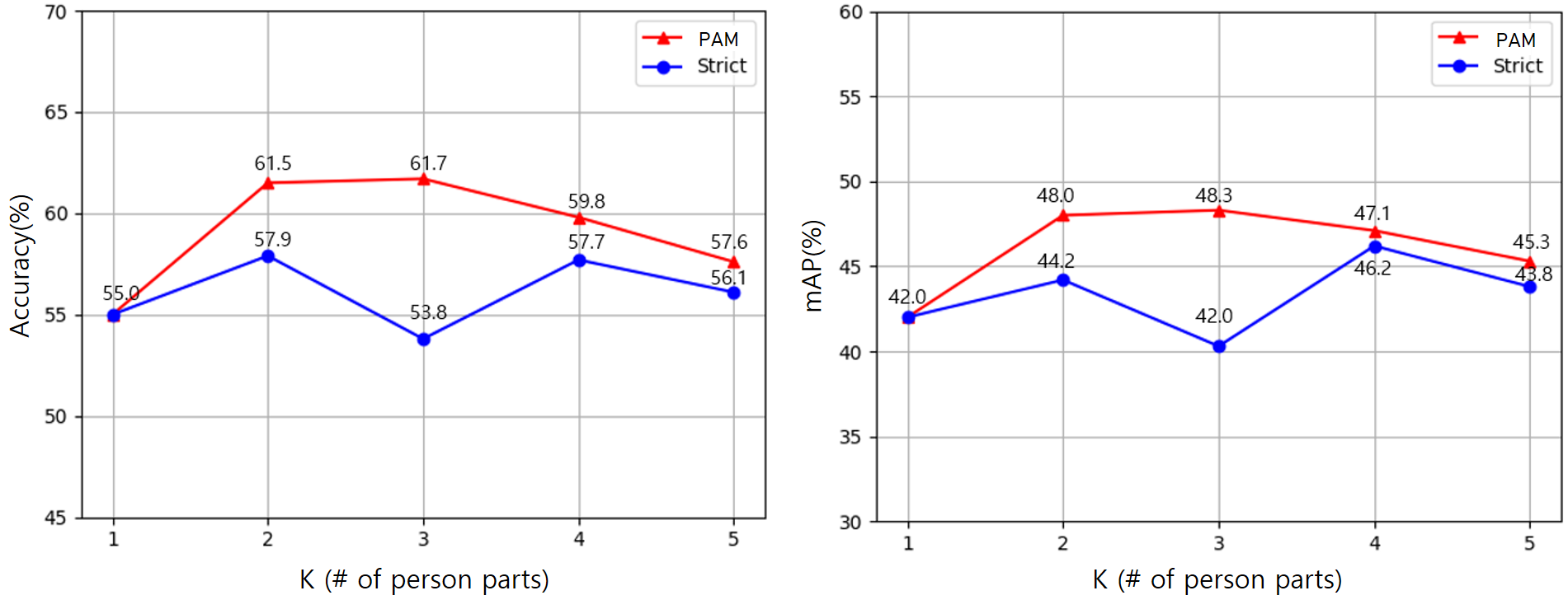}
    \caption{The performance of PADAL$_{mp}$ according to the number of clusters $K$ on the MARS dataset.
    Strict (blue line) denotes the method that leverages uniformly divided features along the vertical line.
    }
     \label{ppavsstrict}
\end{figure}

\textbf{Analysis of a Multi-Domain Discriminator.}
To further validate the importance of domain alignment in person re-identification, we partially aligned the feature distributions between manually selected camera domains.
Specifically, we aligned the ID features across 2- or 3-camera domains by using multi-domain discriminators.
Table \ref{DomainN} shows that the rank-1 accuracy is improved from 42.9 to 55.0 as aligning more camera domains during the network training.
The experimental results show that:
1) There is domain discrepancy between different camera domains, which degrades the performance of the re-ID network.
2) Our MDIFL reduces the camera-domain discrepancy with domain adversarial learning.

\textbf{Visualization of Feature Space.} 
As shown in Fig. \ref{result_tsne}, we visualize the t-SNE of features in the space $F$.
Here, we compare the network with the initialize state, ID-discriminative feature learning, and our PADAL$_{mp}$ method.
The result shows that MDIFL effectively aligns the camera domains.
Furthermore, we visualize the histogram of the feature distance in Fig. \ref{histogram}, which illustrates that our PADAL enhances the ID discrimination ability.

\textbf{The Effect of Cluster Number in PAM.}
In our PAM, we clustered the similar spatial features into $K$ groups by using the K-means clustering algorithm (see Fig. \ref{kpartshow}).
We observe that the number of cluster $K$ affects the performance as shown in Fig. \ref{ppavsstrict}.
The experimental results show that setting $K>3$ degrades the performance, which means that a large cluster number can distract the discrimination ability of the network.
We also experimented on the common approach that uniformly divides the feature map along the vertical line.
The result shows that the adaptive clustering enhances the performance of domain adversarial learning.

\begin{table}[]
\small
\centering
\begin{tabular}{lc|ccc|c}
\Xhline{3\arrayrulewidth}
\multicolumn{2}{c|}{ID Fragment Rate (\%)}               & R1   & R5   & R20  & mAP  \\ \hline
\multicolumn{1}{l|}{\multirow{4}{*}{Baseline}} & 0  & 42.9 & 63.6 & 78.8 & 30.6 \\
\multicolumn{1}{l|}{}                              & 10 &   39.5   & 58.9      &    74.3  & 26.8 \\
\multicolumn{1}{l|}{}                              & 30 &  34.4    &    53.2  & 69.5     &  23.6    \\
\multicolumn{1}{l|}{}                              & 50 &   31.2    &  52.3    &    67.1  & 22.8     \\ \hline
\multicolumn{1}{l|}{\multirow{4}{*}{PADAL$_{mp}$ ($K$=1) }}   & 0  & 55.0 & 72.7 & 84.3 & 42.0 \\
\multicolumn{1}{l|}{}                              & 10 & 52.3 & 72.0 & 83.5 & 39.9 \\
\multicolumn{1}{l|}{}                              & 30 & 51.5 & 71.5 & 83.0 & 39.2 \\
\multicolumn{1}{l|}{}                              & 50 & 50.2 & 70.6 & 82.4 & 38.5\\
\Xhline{3\arrayrulewidth}
\end{tabular}
 \hspace{20mm}
\caption{ Model robustness analysis of ID fragment rates on the MARS dataset.
}
\label{robust}
\end{table}

\textbf{Model Robustness Analysis.}
In real-world situations, the tracklet of one person in a camera can be divided by background clutters and occlusion.
Therefore, the pseudo-labeling method within a camera might contain an ID duplication problem. 
In our experiment, we observe that using domain adversarial learning can reduce the performance degradation from the ID duplication problem.
In Table \ref{robust}, we randomly selected person IDs and divided the tracklet of one person into two tracklets with different IDs.
As the ID fragment rate increases 0\% to 50\%, rank-1 accuracy decreases by 11.7\% (42.9-31.2) and 4.8\% (55.0-50.2), respectively. 
This is because aligning multi-camera domains works as a regularization, where one person's tracklet might be separated in some camera, but not in other cameras.

\section{Conclusion}

In this paper, we have proposed a multi-camera domain feature learning approach for unsupervised person re-identification. 
We maximize feature distances between different person IDs within a camera by using a metric learning approach.
At the same time, we apply domain adversarial learning across multiple camera views for minimizing camera domain discrepancy.
To further enhance Multi-camera Domain Invariant Feature Learning (MDIFL), we suggest the Part-aware Adaptation Module (PAM), which takes the advantage of the spatial information of human body parts.
We carry out comprehensive experiments on several public datasets (i.e., PRID-2011, iLIDS-VID, and MARS) and show the superiority of our PADAL.

{\small
\bibliographystyle{ieee}
\bibliography{egbib}
}

\end{document}